\pdfoutput=1

\documentclass[11pt]{article}

\usepackage[]{acl}

\usepackage{times}
\usepackage{latexsym}

\usepackage[T1]{fontenc}

\usepackage[utf8]{inputenc}

\usepackage{microtype}

\usepackage{inconsolata}

\usepackage{graphicx}

\usepackage{comment}
\usepackage{booktabs}
\usepackage{enumitem}
\usepackage{array}

\newcolumntype{P}[1]{>{\raggedright\arraybackslash}p{#1}}

\title{\includegraphics[height=16px]{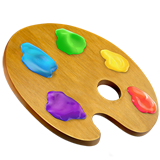} MultimodalHugs: \\ Enabling Sign Language Processing in Hugging Face}

\author{Gerard Sant$^{1}$, Zifan Jiang$^{1}$, Carlos Escolano$^{3}$, Amit Moryossef$^{1, 2}$, \\ \textbf{Mathias Müller}$^{1}$\textbf{,} \textbf{Rico Sennrich}$^{1}$\textbf{,} \textbf{Sarah Ebling}$^{1}$ \\ $^1$University of Zurich, $^2$\href{https://sign.mt}{sign.mt} 
\\ $^3$Barcelona Supercomputing Center
 \\ \texttt{\{gerard.santmuniesa, zifan.jiang, amit.moryossef\}@uzh.ch} \\ \texttt{\{mmueller, sennrich, ebling\}@cl.uzh.ch} \\ \texttt{carlos.escolano@bsc.es}}

\begin{document}
\maketitle
\begin{abstract}

In recent years, sign language processing (SLP) has gained importance in the general field of Natural Language Processing. However, compared to research on spoken\footnote{Following \citet{muller-etal-2022-findings}, we use the word ``spoken'' to refer to any language that is not signed, no matter whether it is represented as text or audio, and no matter whether the discourse is formal (e.g. writing) or informal (e.g. dialogue).} languages, SLP research is hindered by complex ad-hoc code,  inadvertently leading to low reproducibility and unfair comparisons. Existing tools that \textit{are} built for fast and reproducible experimentation, such as {Hugging Face}, are not flexible enough to seamlessly integrate sign language experiments. This view is confirmed by a survey we conducted among SLP researchers.

To address these challenges, 
we introduce \emph{MultimodalHugs}, a framework built on top of Hugging Face that enables more diverse data modalities and tasks, while inheriting the well-known advantages of the Hugging Face ecosystem. Even though sign languages are our primary focus, \emph{MultimodalHugs} adds a layer of abstraction that makes it more widely applicable to other use cases that do not fit one of the standard templates of Hugging Face. We provide quantitative experiments to illustrate how \emph{MultimodalHugs} can accommodate diverse modalities such as pose estimation data for sign languages, or pixel data for text characters.

\end{abstract}

\section{Introduction}

Despite rapid progress in multimodal Natural Language Processing (NLP), sign language processing (SLP) remains an underserved area in terms of tooling and infrastructure. Sign languages are the native languages of millions of deaf individuals worldwide\footnote{According to the \href{https://wfdeaf.org/}{World Federation of the Deaf}.}, and therefore, building mature sign language technologies such as translation systems 
is crucial for inclusion and accessibility \citep{bragg_etal_2019_signlanguagerecognition,yin-etal-2021-including,muller-etal-2022-findings}.

\begin{figure}[ht!]
    \centering
    \includegraphics[width=\columnwidth]{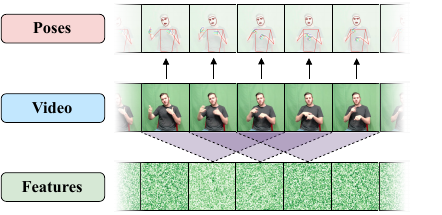}
    \caption{Illustration of the primary modalities used in SLP: extracted skeletal \textit{Poses} highlighting keypoints and joints; raw \textit{Video} frames; and corresponding visual \textit{Features} extracted from videos.}
    \label{fig:modalities}
\end{figure}

While for many spoken languages, technologies are readily available and useful for everyday tasks, that is not the case for sign languages. The reasons include a lack of suitable training data, a smaller research community compared to mainstream NLP, and crucially, a lack of unified, open-source code frameworks to drive research. Current SLP research is characterized by ad-hoc code written for specific experiments with little documentation or proprietary codebases. As a consequence, SLP experiments are hard to reproduce and fair comparisons between papers are challenging (\S\ref{subsec:current-state-of-slp-research}).

Furthermore, SLP systems often involve multiple input types—such as pose sequences (sequences of keypoints representing positions of hands, body, and facial landmarks), RGB videos, and text (Figure~\ref{fig:modalities})—posing unique challenges for multimodal learning \cite{yin-etal-2021-including, tan2024signlanguage, doi:10.1142/S2972335324010038}. This also means that existing NLP frameworks that do emphasize fast and reproducible experimentation, such as Hugging Face \citep{wolf2020transformers}, are not immediately applicable to sign languages (\S\ref{subsec:applicability-of-existing-frameworks}).

To alleviate these issues, we present \emph{MultimodalHugs}, a framework built on top of the Hugging Face ecosystem that extends its capabilities for SLP experiments, for instance by introducing modality-aware processors for non-textual signals.
\emph{MultimodalHugs} remains compatible with Hugging Face’s existing architectures and APIs.

Section~\ref{sec:multimodalhugs} introduces our proposed framework in more detail, and Section~\ref{sec:demonstrations} demonstrates
its capabilities through sign language translation experiments, showcasing its seamless support for diverse multimodal inputs within a unified training pipeline (\S\ref{subsec:slt}). Although our primary focus is on sign language tasks, we show that the same infrastructure generalizes to other non-traditional tasks, such as machine translation from pixel sequences (\S\ref{subsec:pixels}).

Our open-source toolkit is available at \url{https://github.com/GerrySant/multimodalhugs}, and we welcome community contributions to further advance SLP and multimodal research.

\section{Background}
\label{sec:background}

\begin{figure*}[ht!]
    \centering
    \includegraphics[width=\textwidth]{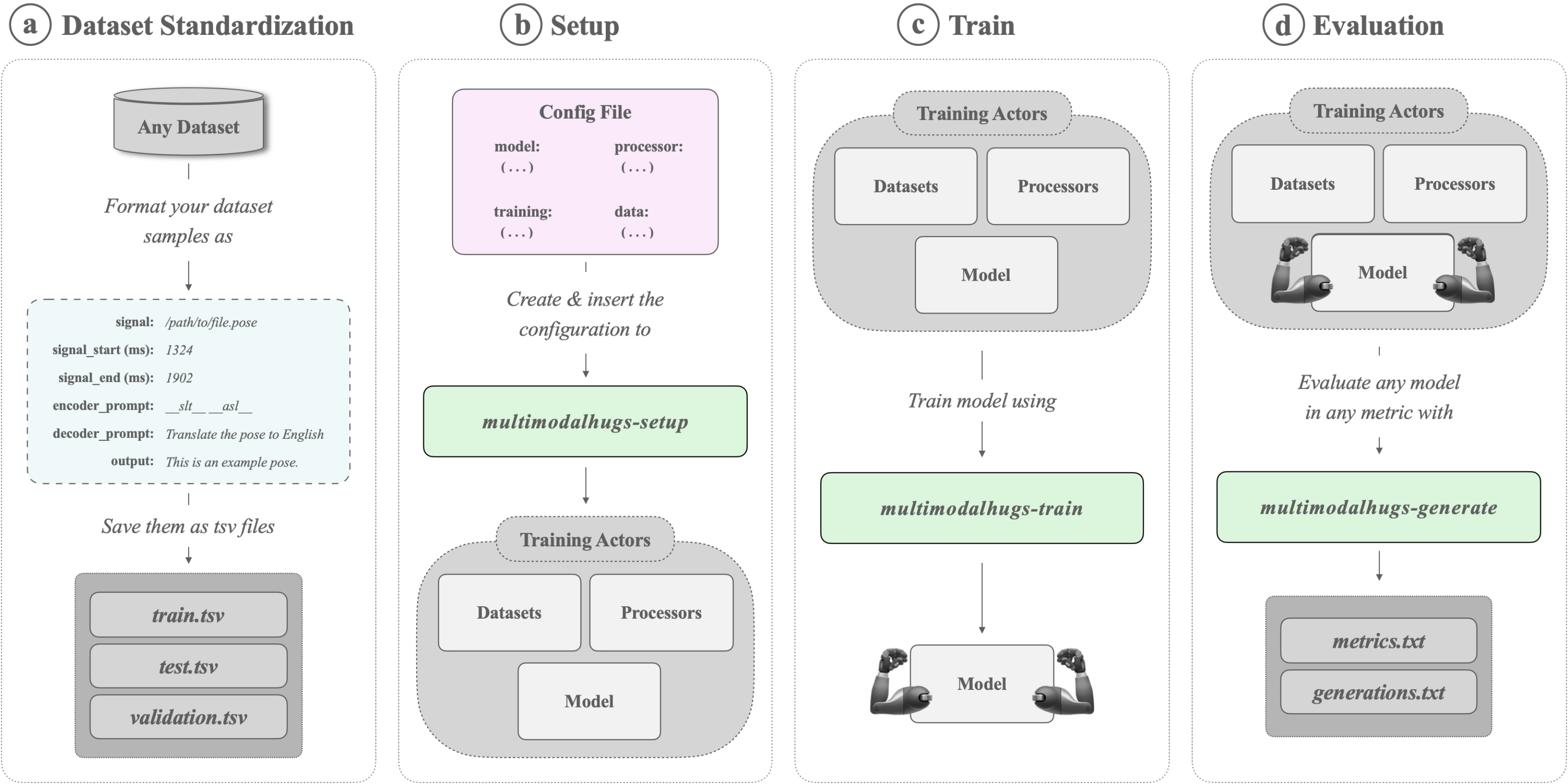}
    \caption{Overview of the \emph{MultimodalHugs} workflow. The figure illustrates the four core steps: a) dataset standardization using TSV files (ASL: American Sign Language); b) setup of training configurations and modules; c) streamlined model training; and d) evaluation through automatic generation and metrics.}
    \label{fig:multimodalhugs-pipeline}
\end{figure*}

\subsection{Current state of SLP research}
\label{subsec:current-state-of-slp-research}

We posit that SLP research currently is held back by a lack of suitable open-source frameworks. As a result, the software engineering part of SLP research is less \textit{sustainable} compared to other sub-fields of NLP and the \textit{scientific validity} of outcomes is sometimes called into question.

\paragraph{Code sustainability}

Existing NLP tools (e.g., Hugging Face Transformers \citep{wolf2020transformers}, OpenNMT \citep{klein-etal-2017-opennmt}, Fairseq \citep{ott-etal-2019-fairseq}) have greatly simplified text-based experimentation but provide little support for SLP experiments.
Therefore, researchers working on SLP often resort to isolated, task-specific pipelines, leading to code duplication, a general sense of starting from scratch for every new experiment, or hard-coded dependencies. In many cases, the code is not available at all. For example, \citet{muller-etal-2023-considerations} surveyed papers on gloss-based sign language translation and found that 7 out of 14 papers did not publish any code.

\paragraph{Scientific validity}

To draw strong conclusions, research aims to be reproducible and enable fair comparisons between papers.
While NLP in general has seen considerable progress in this regard \citep{dodge2020towards,belz-etal-2021-systematic,belz-thomson-2023-2023}, in SLP research it remains a significant challenge. Prior work has highlighted key issues, including inaccessible datasets, poor transparency in preprocessing and evaluation, and a general lack of documentation \citep{waghmare2023signlanguage,muller-etal-2023-considerations}.
Similarly, recent benchmarks such as iSign acknowledge limited baselines and a lack of standard evaluation setups \citep{joshi-etal-2024-isign}.
Broader surveys on LLMs also point to missing prompts, decoding settings, and evaluation scripts as major barriers \citep{laskar-etal-2024-systematic}.

This view is confirmed by an informal survey we conducted among SLP researchers; additional details are provided in Appendix~\ref{sec:appendix_survey}. 
The survey responses indicate that attempting to reproduce SLP experiments is a substantial challenge. Researchers often face undocumented preprocessing steps, unavailable open-source code, and framework incompatibilities. Even when code is shared, adapting it to new modalities or datasets typically requires a lot of effort. Additional barriers include hard-coded paths, dependency conflicts, and restricted access to proprietary models or datasets.

Among respondents who had attempted reproduction, 57.1\% reported that it took over a week, while 14.3\% never succeeded. These responses reflect widespread difficulties in the field.
To improve reproducibility, survey participants emphasized the need for clear documentation, standardized pipelines, and shared environments to ensure structured and reproducible SLP research.

\subsection{Applicability of existing frameworks}
\label{subsec:applicability-of-existing-frameworks}

Existing open-source toolkits, though effective for specific modalities or text-based tasks, provide limited support for general multimodal research and are difficult to extend across domains (see the code sustainability discussion in \S\ref{subsec:current-state-of-slp-research}). Recent vision-language systems like LLaVA \citep{NEURIPS2023_6dcf277e} and Flamingo \citep{NEURIPS2022_960a172b} have achieved strong results in tasks such as visual question answering. However, LLaVA’s open-source version lacks modularity and requires manual setup of data and training scripts, while Flamingo remains closed-source. Even reimplementations like OpenFlamingo \citep{awadalla2023openflamingo} rely on approximations and are hard to adapt. Crucially, none of these frameworks target sign language inputs.

As an illustrative example, consider the widely used \texttt{run\_translation} script from Hugging Face Transformers. This script assumes input and output data as text sequences processed through tokenizers. However, in multimodal research, especially SLP, different modalities—such as pose data, extracted video features, and video sequences—are frequently utilized.

Integrating non-text modalities into this standard Hugging Face pipeline requires significant, task-specific modifications. For instance, numeric pose sequences or video frames cannot directly leverage tokenizer-based preprocessing. Custom preprocessing steps must be developed to convert these diverse modalities into suitable tensor formats. Additionally, data collation, batching, and padding logic must be adapted individually for each modality, quickly becoming cumbersome and highly specific to each experiment. This approach typically results in non-reusable and experiment-specific code.

In contrast, \emph{MultimodalHugs} extends the Hugging Face ecosystem with support for truly multimodal pipelines. Rather than relying on ad-hoc scripts and hard-coded pipelines, researchers can structure their datasets, preprocessing, and model configurations in a unified way, enabling fairer comparisons and reducing duplication of effort. We describe the framework in detail in Section~\ref{sec:multimodalhugs}.

\section{MultimodalHugs} \label{sec:multimodalhugs}

We propose a modular framework that streamlines experimentation for SLP, aimed at addressing the needs of researchers (\S\ref{subsec:current-state-of-slp-research}). It introduces structured components for dataset standardization (\S\ref{subsec:dataset_standardization}), setup (\S\ref{subsec:training_setup}), training (\S\ref{subsec:training}), and evaluation (\S\ref{subsec:evaluation})(Figure~\ref{fig:multimodalhugs-pipeline}), enabling reproducible workflows across varied data modalities. 

The framework builds on the Hugging Face ecosystem and preserves compatibility with its Trainer, tokenizers, and inference APIs.

\subsection{Dataset Standardization} \label{subsec:dataset_standardization}

While Hugging Face supports loading datasets in various formats, integrating new multimodal datasets often requires writing custom scripts or preprocessing logic tailored to each resource. In contrast, \emph{MultimodalHugs} adopts a unified TSV format that standardizes how multimodal data is structured (Figure~\ref{fig:multimodalhugs-pipeline}a). Preprocessing is decoupled from datasets and handled separately by modality-specific processors (\S\ref{subsec:training_setup}). Each TSV row defines one example and includes the following fields:

\begin{itemize}[left=4pt]
    \item \texttt{signal}: the primary input, either as raw text or a path to a multimodal resource (e.g., image, pose sequence\footnote{Pose refers to keypoint-based representations of human body movements~\cite{10.1145/3603618}.}, or audio).
    \item \texttt{signal\_start} and \texttt{signal\_end}: optional temporal markers for time-based inputs (clips).
    \item \mbox{\texttt{encoder\_prompt} and \texttt{decoder\_prompt}:} optional text fields used to condition the input or output; useful for injecting task instructions, modality tags, or language cues.\footnotemark
    \item \texttt{output}: the target output.
\end{itemize}

\footnotetext{Prompt fields may be left empty if the model lacks an encoder or decoder.}

Each dataset split (train, validation, test) is provided as a separate TSV file. This standardized interface simplifies the integration of multimodal data into training pipelines and enables straightforward per-sample customization through fine-grained control over prompts or clipping—features that would otherwise require manual preprocessing logic. Appendix~\ref{appendix:scaling-slt-reproduction} shows how \emph{MultimodalHugs} simply reproduction of a complex state-of-the-art SLP system without additional code, leveraging the potential of the proposed dataset format.

\subsection{Setup} \label{subsec:training_setup}

The setup process is responsible for initializing the core components of the training pipeline, referred to as \textbf{Training Actors}: Datasets, Processors, and Model. These actors are instantiated based on a user-defined configuration, ensuring that multimodal inputs are properly structured, preprocessed, and integrated into a compatible model. An example illustrating the interaction between these components is provided in Appendix~\ref{appendix:pose-workflow}.

\begin{itemize}[left=4pt]
    \item \textbf{Datasets}: The dataset actor structures TSV inputs into a format suitable for training—such as a Hugging Face Dataset or a PyTorch Dataset—supporting batching, shuffling, and efficient loading. It abstracts away the underlying file structure, allowing seamless integration with training components.

    \item \textbf{Processors}: The processor actor transforms each dataset sample into the input format expected by the model. Unlike Hugging Face pipelines, where tokenizers and processors are often coupled to specific architectures or tasks, processors in \emph{MultimodalHugs} are designed to be modular and model-agnostic. This separation enables reusability across datasets and tasks while maintaining support for modality-specific operations such as tokenization, feature extraction, and sequence alignment.

    \item \textbf{Model}: Users can define their own models or use any of the default customizable architectures provided by \emph{MultimodalHugs}. Default models usually consist of three primary components: a \textbf{Feature Extractor} that processes raw multimodal signals into numerical representations, a \textbf{Multimodal Mapper} that aligns extracted features with text-based embeddings, and a \textbf{Backbone} that generates output text conditioned on the multimodal input. Users can modify or replace any of these components, integrate their own architectures, or directly use pretrained models from Hugging Face.
\end{itemize}

Custom datasets, processors, and models can be easily implemented following the framework’s documentation.

\paragraph{Configuration File.} A YAML file defines all parameters required to initialize the training environment. It is organized into four sections: 

(i) \textbf{Model} specifies architectural hyperparameters, such as transformer backbone and embedding details, among others;
(ii) \textbf{Data} describes settings related to the dataset, including paths to standardized TSV files, existing dataset instances or data filtering criteria.
(iii) \textbf{Processor} includes any arguments needed to configure the desired preprocessing pipeline; and
(iv) \textbf{Training} defines the training hyperparameters and optimization settings leveraging Hugging Face's \texttt{Trainer} class.
This configuration file ensures that training runs are reproducible and customizable. Users can refer to the documentation for a full list of configurable arguments.

\paragraph{Automatic Setup.} Once the configuration file is defined, the training actors are instantiated with:
\begin{lstlisting}
multimodalhugs-setup \
    --modality <modality_name> \
    --config_path <config_path>
\end{lstlisting}

This command ensures that all necessary components—datasets, processors, and models—are correctly instantiated and linked, eliminating the need for manual setup.

\subsection{Training} \label{subsec:training}

The training pipeline is built on Hugging Face’s \texttt{Trainer} API, allowing seamless integration with existing workflows, including support for mixed-precision training, distributed learning, and flexible optimization strategies. Once the configuration is defined (\S~\ref{subsec:training_setup}), training can be launched with:

\begin{lstlisting}
multimodalhugs-train --task <task> \
    [--training_args]
\end{lstlisting}

This command automatically instantiates the necessary components—datasets, processors, and model—executes the training loop, and manages checkpointing and evaluation.

\begin{figure*}[ht!]
    \centering
    \includegraphics[width=\textwidth]{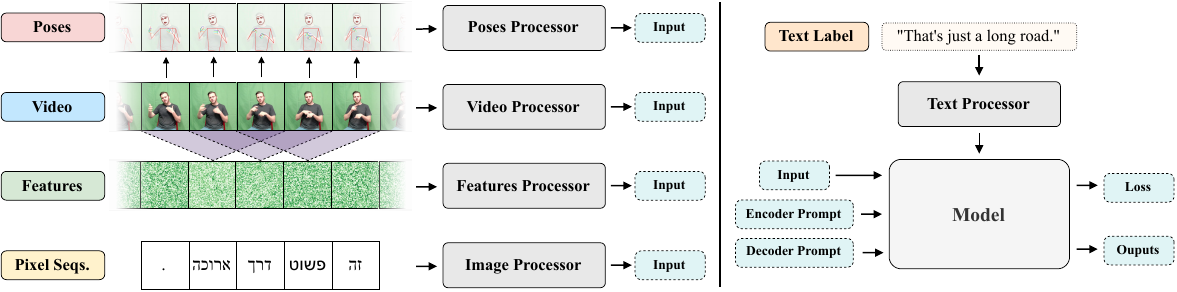}
    \caption{Schematic overview of the \emph{MultimodalHugs} processing pipeline, covering all experiments presented in both demonstrations. Sequences can come from different modalities, each processed by a dedicated processor to produce the appropriate input tensors for the model. The right side illustrates the translation training workflow, where text labels are also processed and provided as targets.}
    \label{fig:experiments_pipelines}
\end{figure*}

\subsection{Evaluation} \label{subsec:evaluation}

Evaluation is performed using the \texttt{generate} command, which loads the trained model and associated components, runs inference on the test split, and computes the specified evaluation metrics:

\begin{lstlisting}
multimodalhugs-generate --task <task> \
    --metric_name <metric_name> \
    --config_path <config_path> \
    [--other_args]
\end{lstlisting}

The framework supports standard metrics (e.g., BLEU \citep{papineni-etal-2002-bleu}, COMET \cite{rei-etal-2020-comet}, perplexity \citep{10.1121/1.2016299}) via Hugging Face’s \texttt{evaluate} library \citep{von-werra-etal-2022-evaluate}. Predictions are decoded, postprocessed, and saved along with reference labels to facilitate transparent analysis (example shown in Appendix~\ref{sec:appendix_predictions}). 

\footnotetext{As in training, paths to Training Actors instances can be specified either through the YAML configuration or directly from the command line.}

\section{Demonstrations}\label{sec:demonstrations}

We demonstrate the ease of use and flexibility of \emph{MultimodalHugs} through two multimodal translation tasks (Figure~\ref{fig:experiments_pipelines}). We first highlight support for sign language processing through pose- and video-based models, then extend the same infrastructure to a pixel-based translation task—demonstrating rapid experimentation across modalities and architectures with minimal adjustments.

\subsection{Sign Language Translation}  \label{subsec:slt}

Sign language translation involves converting between sign language videos and spoken or written language. Despite recent advances, it remains a challenging multimodal task \citep{Camgoz_2020_CVPR, NEURIPS2022_6cd3ac24, de-coster-etal-2021-frozen, 1360016867511293824}. A common approach for sign-to-text translation is to extract pose estimations as video-level features \citep{openpose, mediapipe, 9667087, muller-etal-2022-findings, NEURIPS2023_5c61452d, zhang2024scalingsignlanguagetranslation}.

We investigate whether RGB video features outperform MediaPipe \citep{mediapipe} pose representations for translation. As a baseline, we replicate a pose-to-text system from \citet{NEURIPS2023_5c61452d}, which uses a linear projection layer followed by an \texttt{mT5-base} model \citep{xue2021mt5massivelymultilingualpretrained}, trained on the How2Sign dataset \mbox{\citep{Duarte_2021_CVPR}.}

Using \emph{MultimodalHugs}, we reproduce their results by (i) organizing the dataset in the TSV format required by the framework, (ii) selecting a linear layer as the Multimodal Mapper and \texttt{mT5} as the backbone in the configuration file, and (iii) specifying the desired modality and task when running the setup and training commands. This setup replicates the original system (Table~\ref{tab:ASL_translation}, where row 0 reports the result from \citet{NEURIPS2023_5c61452d} and row 1 our reproduction) without requiring any additional code.

\begin{table}[t]
    \centering
    \resizebox{\columnwidth}{!}{%
    \begin{tabular}{lllrr}
        \hline
        \textbf{Nº} & \textbf{Approach} & \textbf{Modality} & \textbf{chrF} & \textbf{BLEU} \\
        \hline
        0 & \citet{NEURIPS2023_5c61452d} & pose & - & 0.86 \\
        \hline
        1 & Linear $+$ \texttt{mT5-base} & pose & 18.14 & 1.43 \\
        2 & Linear $+$ \texttt{M2M-100} & pose & 23.71 & 4.98 \\
        \hline
        3 & \texttt{I3D} $+$ \texttt{mT5-base} & features & 24.78 & 5.21 \\
        4 & \texttt{I3D} $+$ \texttt{M2M-100} & features & 30.16 & 9.19 \\
        \hline
        5 & Linear $+$ \texttt{mT5-base} & video & 23.52 & 4.30 \\
        6 & Linear $+$ \texttt{M2M-100} & video & 27.43 & 7.82 \\
        \hline
    \end{tabular}%
    }
    \caption{Sign language translation results. Models are evaluated using chrF \citep{popovic-2015-chrf} and BLEU \citep{papineni-etal-2002-bleu} scores, comparing pose-based and video-based methods.}
    \label{tab:ASL_translation}
\end{table}

We then replace the pose‐based inputs with pre‐computed \texttt{I3D} features \citep{Carreira_2017_CVPR} from RGB frames, following the approach of \citet{Tarres_2023_CVPR}, and also explore an end-to-end video-to-text setup where raw RGB frames are fed directly into the linear mapper (modality set to \texttt{video}). Leveraging the same training pipeline, we switch to the appropriate processor for each modality and update dataset pointers to reference the new input files (e.g., from \texttt{.pose} to \texttt{.mp4}). These minimal adjustments allow us to train both a feature-based and an end-to-end video translation model. As shown in Table~\ref{tab:ASL_translation}, using pre-computed \texttt{I3D} features (row 3) improves performance over the pose-based baseline, while the end-to-end video approach (row 5) simplifies the pipeline but yields slightly lower results compared to the \texttt{I3D} variants.

To validate whether these gains are consistent across language models, we repeat all experiments using \texttt{M2M-100} \citep{m2m} as the backbone. By changing the \texttt{backbone\_type} in the configuration file, we obtain the results shown in Table \ref{tab:ASL_translation} (rows 2, 4 and 6). We observe that both video-feature (row 4) and end-to-end video (row 6) models outperform MediaPipe pose-based models, regardless of the backbone used, which coincides with the findings of \citet{muller2023d4} who performed similar experiments.

\subsection{Machine Translation From Pixels} \label{subsec:pixels}

For some languages, standard tokenization techniques are not well-suited \cite{tsarfaty-etal-2010-statistical, klein-tsarfaty-2020-getting,info12020052, singh-etal-2023-subwords}. These include languages with \mbox{complex} scripts that are composed using multiple glyphs (e.g. Hangul, Tamil or Telugu); writing systems such as SignWriting \cite{sutton1990lessons} for signed languages; and standard scripts for highly morphologically rich languages like Hebrew \cite{doi:10.1287/ijds.2022.0016, seker-etal-2022-alephbert}, Arabic \cite{ALAZANI2017359, 8717369, fouad2020arwordvec, faris2020hate}, and Turkish \cite{kuriyozov-etal-2020-cross,AYDOGAN2020123288}.

We examine the case of Hebrew, a root-and-template language. While existing translation models support it, they require roughly \textbf{16\% more} tokens than their English counterparts, and splitting up words into subtokens may not properly represent the language. Therefore, after pre-tokenization (whitespace/punctuation splits) and inspired by \citet{DBLP:conf/iclr/RustLBSLE23}, which models language from visual renderings of text, we experiment with encoding each `word' as an image of that word.
This method bypasses the NP-complete tokenization challenge \cite{whittington2024tokenisationnpcomplete} by compressing information visually.

We use our framework to fine tune \texttt{M2M-100} \cite{m2m} on the Hebrew$\rightarrow$English NLLB dataset \citep{schwenk2020ccmatrixminingbillionshighquality, nllbteam2022languageleftbehindscaling}, by writing a processor that takes each sentence, splits it into `words' and returns a sequence of images. The framework treats the sequence as a video, and the model embeds each `frame' using \texttt{CLIP} \cite{clip}. During training, the model is kept completely frozen; only the shared embedding matrix—used both for input embeddings and the output projection—is fine-tuned.

\begin{table}[ht]
    \centering
    \resizebox{\columnwidth}{!}{%
    \begin{tabular}{lcccc}
        \hline
        \textbf{Model} & \textbf{chrF} & \textbf{BLEU} &\textbf{XCOMET-XXL} \\
        \hline
        \texttt{M2M-100} & 58.89 & 31.81 & 75.92 \\
        \texttt{CLIP} + \texttt{M2M-100} & 58.63 & 31.66 & 75.15 \\
        \hline
    \end{tabular}%
    }
    \caption{Hebrew-to-English translation results.}
    \label{tab:he_en_results}
\end{table}

With our framework, we set up this experiment without modifying the model code or dealing with additional training pipeline complexity.
Table \ref{tab:he_en_results} shows that our new Hebrew-to-English translation model performs on par with the original model based on automatic metrics, while requiring \textbf{45\% fewer} tokens, which is useful in limited memory or large-context use cases.
A manual evaluation by a native Hebrew speaker indicates that our image-based model produces translations judged as \textcolor{Green}{46\% better} translations, \textcolor{orange}{30\% on-par}, and only \textcolor{Red}{24\% worse} (Appendix \ref{appendix:hebrew})
compared to the baseline. Notably, our approach excels at translating named entities and misspellings, while the baseline remains superior at copying words verbatim.

\section{Conclusions and Future Work} \label{sec:conclusion_and_future_work}

We presented \emph{MultimodalHugs}, a framework developed to support sign language processing within the Hugging Face ecosystem. 
It introduces general-purpose features—such as modality-specific processors, a standardized dataset interface, and a flexible training pipeline—that also simplify experimentation in other multimodal tasks. These components, among other design choices, reduce entry barriers and promote reproducible research in underrepresented domains.

We demonstrate how standard sign language pipelines can be replicated and extended with minimal effort, and how the same infrastructure generalizes to non-traditional use cases, such as visual-language modeling without the need for tokenizers. 

Inspired by decoder-only models like LLaVA \cite{NEURIPS2023_6dcf277e}, we plan to support mixed-modality input sequences in future work, allowing users to interleave textual and non-textual signals freely. This could be achieved through a new workflow (see Appendix~\ref{appendix:future-workflow}) that automatically detects, embeds, and aligns multimodal signals, giving full control over input composition and placement.

\section*{Ethics Statement}

When working with sign language data special care needs to be taken to protect the privacy of individuals. We do not advocate research on models that generate a person's likeness, such as for instance their face. Similarly, we do not condone research using data without express permission by the people shown in the sign language videos. Even representations derived from videos such as poses may retain some identifying features \citep{battisti-etal-2024-person}, i.e. are not anonymous and the same considerations as for videos apply to such derivates. The How2Sign sign language data \citep{Duarte_2021_CVPR} we use in Section \ref{subsec:slt} is distributed under a Creative Commons Attribution-NonCommercial 4.0 International License and we are confident that we are using this dataset as intended by the license terms.


\clearpage
\appendix

\section{Survey} \label{sec:appendix_survey}

To better understand the challenges in multimodal and sign language processing and assess the usability of \emph{MultimodalHugs}, we conducted a survey (more details in Appendix~\ref{sec:raw_appendix_survey}) among researchers in the field. The questionnaire addressed reproduction difficulties, framework preferences, and usability feedback.

\subsection{Survey Questionnaire and Responses} \label{sec:raw_appendix_survey}

To provide transparency on the survey methodology, we include the full list of questions and a summary of responses. The survey was conducted among researchers in multimodal and sign language processing to assess challenges in reproducibility and evaluate the usability of \mbox{\emph{MultimodalHugs}.}

\subsection*{Survey Questions}

The survey consisted of three main sections: Background Information, Reproducibility Challenges, and \emph{MultimodalHugs} Evaluation.

\begin{enumerate}[left=4pt]
    \item \textbf{Background Information}
        \begin{itemize}[left=-4pt]
            \item Years of experience in multimodal or sign language research (Less than 1 year, 1--3 years, 3--5 years, More than 5 years)
            \item Primary area of research
            \item Highest level of education (Bachelor’s, Master’s, PhD)
        \end{itemize}
        
    \item \textbf{Reproducibility Challenges}
        \begin{itemize}[left=-4pt]
            \item Have you attempted to reproduce a multimodal/sign language experiment? (Yes/No)
            \item If yes, which paper or project?
            \item What challenges did you face? (Multiple selections allowed)
            \item Time taken to replicate the experiment
            \item What would have made replication easier? (Open-ended)
        \end{itemize}

    \item \textbf{\emph{MultimodalHugs} Evaluation}
        \begin{itemize}[left=-4pt]
            \item Rate ease of use (Scale: 1--5)
            \item Preferred data preparation format (TSV, custom loader, etc.)
            \item Most valuable component of the framework
            \item Willingness to use it as primary framework in the future (Yes/No)
            \item Desired features or suggestions (Open-ended)
        \end{itemize}
\end{enumerate}
\subsection*{Summary of Responses}

A total of 16 valid responses were collected.

\paragraph{Participant Background.}
\begin{itemize}[left=4pt]
    \item \textbf{Experience:} 6.25\% had less than 1 year, 43.75\% had 1--3 years, 31.25\% had 3--5 years, and 18.75\% had more than 5 years of \mbox{experience.}
    \item \textbf{Education:} 6.25\% held a Bachelor's, 62.5\% held a Master’s, and 31.25\% held a PhD.
    \item \textbf{Domains:} 75\% were in sign language-related fields (translation, recognition, or generation); the rest worked in related areas such as semantic representations, visual content generation, interpretability, medical image analysis, and remote sensing.
\end{itemize}

\paragraph{Reproducibility Challenges.}
\begin{itemize}[left=4pt]
    \item \textbf{Reproduction Attempts:} 87.5\% (14 out of 16) had attempted to reproduce a multimodal/sign language experiment.
    \item \textbf{Main Issues:}
    \begin{itemize}[left=-4pt]
        \item Coding challenges: 85.7\%
        \item Undocumented preprocessing steps: 71.4\%
        \item Lack of support for specific modalities: 42.9\%
        \item Proprietary models or datasets: 35.7\%
        \item Missing dataset details: 21.4\%
        \item Inconsistent label filtering/modification: 7.1\%
    \end{itemize}
    \item \textbf{Time to Reproduce:}
    \begin{itemize}[left=-4pt]
        \item More than a week: 57.1\%
        \item About a week: 14.3\%
        \item 1--3 days: 7.1\%
        \item Never succeeded: 14.3\%
        \item Not applicable (no attempt): 7.1\%
    \end{itemize}
\end{itemize}

\paragraph{Ease of Use.}
\begin{itemize}[left=4pt]
    \item \textbf{Rating:} 25\% rated setup as 5 (extremely easy), 62.5\% as 4, 6.25\% as 3, and 6.25\% as 2.
\end{itemize}

\paragraph{Dataset Preparation.}
\begin{itemize}[left=4pt]
    \item \textbf{Preferred Format:} 68.75\% preferred TSV-based dataset standardization; 31.25\% preferred a custom data loader or had no strong preference.
\end{itemize}

\paragraph{Framework Features.}
\begin{itemize}[left=4pt]
    \item \textbf{Most Valuable Features:}
    \begin{itemize}[left=-4pt]
        \item Standardized dataset preparation: 81.25\%
        \item Modular training pipeline: 68.75\%
        \item Flexible integration w/ Hugging Face: 68.75\%
        \item Comprehensive documentation and modular design: 56.25\%
    \end{itemize}
\end{itemize}

\paragraph{Adoption Intent.}
\begin{itemize}[left=4pt]
    \item 87.5\% (14 out of 16) would consider using \emph{MultimodalHugs} as their primary framework in future multimodal experiments.
    \item Reasons for hesitation included early-stage maturity, limited testing, and unclear pip installability.
\end{itemize}

\paragraph{Suggested Improvements.}
Respondents suggested the following additions or improvements:
\begin{itemize}[left=4pt]
    \item Google Colab notebook with a “run all” demo (18.75\%)
    \item Clear support for multiple simultaneous input modalities (e.g., pose + video, text + gloss)
    \item Profiling tools (e.g., to track preprocessing vs. model time)
    \item Support for multimodal targets and non-text outputs
    \item Interactive demos and dry-run examples
    \item More flexibility in dataset schema (e.g., adding gloss or non-manual signals)
    \item Publishing preprocessed datasets through Hugging Face Datasets
\end{itemize}

\subsection{Summary of Survey Results} \label{sec:survey_results}
Participants in the survey represented a wide range of experience levels and academic backgrounds. In terms of research experience, 6.25\% had less than 1 year, 43.75\% had 1--3 years, 31.25\% had 3--5 years, and 18.75\% had over 5 years. Most held a Master’s degree (62.5\%), followed by PhDs (31.25\%) and Bachelor’s degrees (6.25\%). A majority (75\%) specialized in sign language processing, while others worked in related fields such as semantic representations, computer vision, interpretability, medical imaging, and remote sensing.

A total of 87.5\% had tried to reproduce multimodal/sign language experiments. Reported challenges included coding difficulties (85.7\%), undocumented preprocessing (71.4\%), lack of modality support (42.9\%), proprietary models/data (35.7\%), and missing dataset details (21.4\%).

\emph{MultimodalHugs} was generally rated easy to use by most respondents (87.5\% gave a rating of 4 or 5 on a 1–5 scale, with 5 as "extremely easy"). For dataset preparation, 68.75\% preferred the TSV-based format, while 31.25\% favored custom loaders or had no preference. Despite concerns about TSVs for complex data, the structure was appreciated. Valued features of \emph{MultimodalHugs} included dataset standardization (81.25\%), modular training (68.75\%), Hugging Face integration (68.75\%), and documentation/modularity (56.25\%).

Suggested improvements included support for more tasks (e.g., video-to-text), profiling tools, multiple multimodal input support (e.g., pose + video), and an interactive Colab demo (18.75\%).

Overall, the survey highlights strong interest in \emph{MultimodalHugs} as a reproducibility-focused multimodal framework, with 87.5\% of respondents considering to use it as their main framework. The valuable insights gathered will guide \mbox{future development.}

\onecolumn

\section{Training Workflow Example for Pose Sequences}
\label{appendix:pose-workflow}

\begin{figure*}[ht!]
    \centering
    \includegraphics[width=\textwidth]{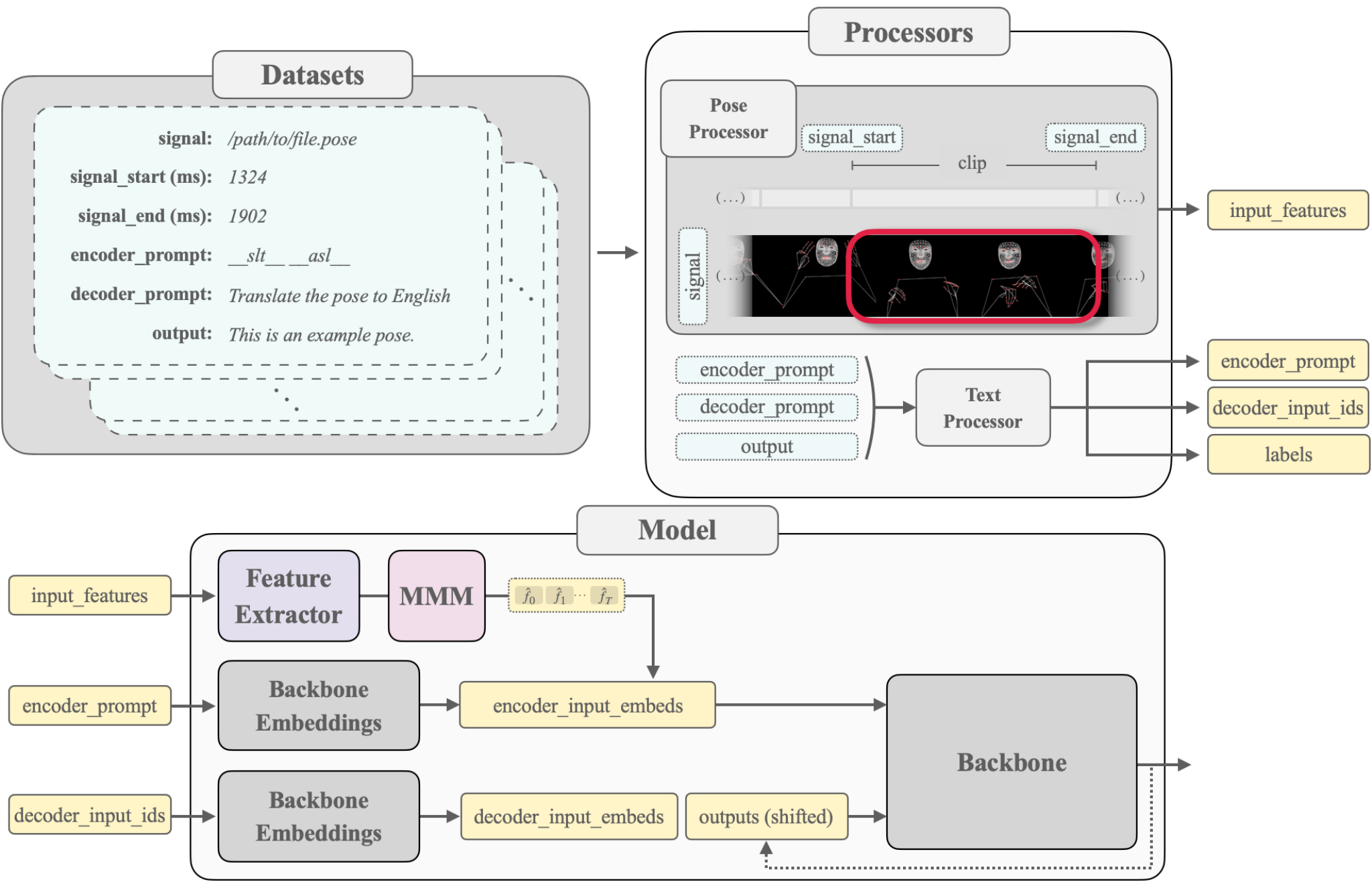}
    \caption{Example training pipeline for a pose-based experiment in \emph{MultimodalHugs}. Datasets are structured into a standardized TSV format and processed by a pose-specific processor. The processor outputs are converted into model-ready inputs: multimodal features, prompts, and labels. The model combines feature extraction, a Multimodal Mapper (MMM), and a language model backbone to generate outputs. The pose frame images are adapted from \citet{zhang2024scalingsignlanguagetranslation}.}
    \label{fig:multimodalhugs_current_training_workflow}
\end{figure*}

Figure~\ref{fig:multimodalhugs_current_training_workflow} shows an example of the workflow between the different training actors in \emph{MultimodalHugs}, specifically for a pose-based experiment. This diagram illustrates the end-to-end process used in our first demonstration setup: pose sequences are loaded from standardized TSV files, preprocessed using a modality-specific processor, and transformed into model-ready inputs. The model integrates a feature extractor, a Multimodal Mapper (MMM), and a language model backbone to produce the final output.

\clearpage

\section{Reproducing the SOTA in Sign Language Translation}
\label{appendix:scaling-slt-reproduction}

\begin{figure}[h]
    \centering
    \includegraphics[width=0.75\columnwidth]{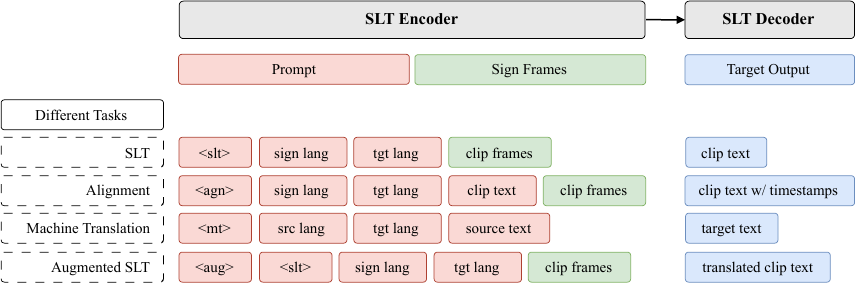}
    \caption{Multitask strategy proposed by \citet{zhang2024scalingsignlanguagetranslation}. The figure illustrates an encoder-decoder SLT model trained on four pretraining tasks, all implemented as sequence-to-sequence learning within a shared architecture. Input prompts (\textcolor{red}{red}), sign frames (\textcolor{Green}{green}), and target outputs (\textcolor{blue}{blue}) are combined in task-specific ways: ``\texttt{sign lang}'' refers to the sign language name; ``\texttt{src lang}/\texttt{tgt lang}'' are source and target spoken languages; ``\texttt{<}\mbox{\texttt{*}}\texttt{>}'' are task control tokens (e.g., \texttt{<slt>}, \texttt{<agn>}, \texttt{<aug>}, \texttt{<mt>}); ``\texttt{source}/\texttt{target text}'' are sentences for MT; ``\texttt{clip frames} (\texttt{clip text})'' denote pose sequences and their spoken captions for a video clip; ``\texttt{translated clip text}'' refers to augmented data obtained via external MT; and ``\texttt{clip text w/ timestamps}'' provides alignment of captions with temporal markers.}

    \label{fig:scaling-multitask}
\end{figure}

This appendix shows how \emph{MultimodalHugs} enables the reproduction of current state-of-the-art systems in Sign Language Translation (SLT) with minimal effort. In particular, we describe the steps required to replicate the multitask system proposed by \citet{zhang2024scalingsignlanguagetranslation}, which combines SLT, alignment, machine translation (MT), and augmented SLT within a unified encoder–decoder framework based on ByT5 \citep{DBLP:journals/tacl/XueBCANKRR22}. Using standardized TSV inputs and a YAML configuration file, \emph{MultimodalHugs} supports this setup without the need for any custom code. Figure~\ref{fig:scaling-multitask} illustrates the multitask training strategy.

\vspace{-2.5ex}
\subsection{Standardized Multitask Dataset TSV Creation}

Each of the tasks described by \citet{zhang2024scalingsignlanguagetranslation} can be expressed using our standardized TSV format, which enables multimodal input and prompt-driven control via several fields. Below we describe the structure for each task.

\paragraph{Sign Language Translation (SLT).} The authors use YouTube SLT videos \citep{tanzer2024youtubesl25largescaleopendomainmultilingual} and apply pose estimation using MediaPipe Holistic \citep{mediapipe, grishchenko2020mediapipe},  which can efficiently be estimated using the Python package from \citet{DBLP:journals/corr/abs-2310-09066}. After the previous step, pose files (e.g., in \texttt{.pose} or \texttt{.json} format) are obtained. Dataset metadata per sample must be then organized following the \emph{MultimodalHugs} format and structured specifically for the SLT task as shown in the table~\ref{tab:slt}.

\begin{table*}[ht]
\centering
\renewcommand{\arraystretch}{1.2}
\resizebox{\textwidth}{!}{
\begin{tabular}{|p{3.65cm}|p{2.5cm}|p{2cm}|p{4.5cm}|p{2.7cm}|p{12cm}|}
\hline
\textbf{signal} & \textbf{signal\_start} & \textbf{signal\_end} & \textbf{encoder\_prompt} & \textbf{decoder\_prompt} & \textbf{output} \\
\hline
\texttt{/path/to/pose1.pose} & \texttt{signal\_start} & \texttt{signal\_end} & \texttt{<slt> sign\_lang tgt\_lang} &  & \texttt{clip\_text} \\
\hline
\texttt{/path/to/pose2.pose} & \texttt{404} & \texttt{514} & \texttt{<slt> asl en} &  & \texttt{Moving the stick adjusts the wing’s angle of attack.} \\
\hline
\texttt{/path/to/pose3.pose} & \texttt{63} & \texttt{88} & \texttt{<slt> csl es} &  & \texttt{El elevador ajusta el ángulo de ataque del avión.} \\
\hline
\end{tabular}
}
\caption{TSV structure for Sign Language Translation.}
\label{tab:slt}
\end{table*}

\vspace{-2.5ex}
\paragraph{Alignment.}
This task can reuse SLT videos or introduce new ones. Multiple SLT clips from the same video are grouped into larger meta-segments using their temporal boundaries (as shown in table~\ref{tab:alignment}). The prompt includes the spoken transcript and the model is trained to predict sentence-aligned time spans.

\begin{table*}[!h]
\centering
\renewcommand{\arraystretch}{1.2}
\resizebox{\textwidth}{!}{
\begin{tabular}{|p{3.65cm}|p{2.5cm}|p{2cm}|p{5.5cm}|p{2.7cm}|p{11cm}|}
\hline
\textbf{signal} & \textbf{signal\_start} & \textbf{signal\_end} & \textbf{encoder\_prompt} & \textbf{decoder\_prompt} & \textbf{output} \\
\hline
\texttt{/path/to/pose1.pose} & \texttt{sample\_start} & \texttt{sample\_end} & \texttt{<agn> sign\_lang tgt\_lang clip\_text} &  & \texttt{clip\_text\_with\_time\_stamps} \\
\hline
\texttt{/path/to/pose2.pose} & \texttt{0} & \texttt{4000} & \texttt{<agn> asl en Hello everyone. Today’s weather...} &  & \texttt{[00:00–00:02] Hello everyone. [00:02–00:04] Today’s weather is sunny...} \\
\hline
\texttt{/path/to/pose3.pose} & \texttt{63000} & \texttt{88000} & \texttt{<agn> csl es Me llamo Elisenda. Estudié...} &  & \texttt{[00:01:03–00:01:10] Me llamo Elisenda. [00:01:10–00:01:28] Estudié...} \\
\hline
\end{tabular}
}
\caption{TSV structure for Alignment Task.}
\label{tab:alignment}
\end{table*}

\paragraph{Machine Translation.}
For the MT task, the authors use a subset of the MADLAD-400 dataset \citep{NEURIPS2023_d49042a5}. Since this task does not involve any multimodal signal, only text is used as input and output. Consequently, the \texttt{signal}, \texttt{signal\_start}, and \texttt{signal\_end} fields in the TSV remain empty. The source sentence, together with the task control token and language pair, is placed in the \texttt{encoder\_prompt}, and the target sentence is placed in the \texttt{output} field, as illustrated in the table below.

\begin{table*}[ht]
\centering
\renewcommand{\arraystretch}{1.2}
\resizebox{\textwidth}{!}{
\begin{tabular}{|p{1.5cm}|p{2cm}|p{2cm}|p{9.15cm}|p{2.7cm}|p{10cm}|}
\hline
\textbf{signal} & \textbf{signal\_start} & \textbf{signal\_end} & \textbf{encoder\_prompt} & \textbf{decoder\_prompt} & \textbf{output} \\
\hline
 &  &  & \texttt{<mt> src\_lang tgt\_lang source\_text} &  & \texttt{target\_text} \\
\hline
 &  &  & \texttt{<mt> es en El gato se sienta en la estera.} &  & \texttt{The cat sits on the mat.} \\
\hline
 &  &  & \texttt{<mt> fr de Je vous remercie pour votre aide.} &  & \texttt{Vielen Dank für Ihre Hilfe.} \\
\hline
\end{tabular}
}
\caption{TSV structure for Machine Translation.}
\label{tab:mt}
\end{table*}
\vspace{-1.5ex}

\paragraph{Augmented SLT.}
To generate augmented SLT, the spoken translation in each SLT example is translated into additional languages (e.g., Italian, Basque) using an external MT system. These translations are then added as new examples, preserving the same pose input. As shown in table~\ref{tab:aug_slt}, the special \texttt{<aug>} token is added into the \texttt{encoder\_prompt}.

\begin{table*}[ht]
\centering
\renewcommand{\arraystretch}{1.2}
\resizebox{\textwidth}{!}{
\begin{tabular}{|p{3.65cm}|p{2.5cm}|p{2cm}|p{7cm}|p{2.7cm}|p{9.5cm}|}
\hline
\textbf{signal} & \textbf{signal\_start} & \textbf{signal\_end} & \textbf{encoder\_prompt} & \textbf{decoder\_prompt} & \textbf{output} \\
\hline
\texttt{/path/to/pose1.pose} & \texttt{signal\_start} & \texttt{signal\_end} & \texttt{<aug> <slt> sign\_lang tgt\_lang} &  & \texttt{augmented\_clip\_text} \\
\hline
\texttt{/path/to/pose2.pose} & \texttt{404} & \texttt{514} & \texttt{<aug> <slt> asl it} &  & \texttt{Spostando il timone...} \\
\hline
\texttt{/path/to/pose3.pose} & \texttt{63} & \texttt{88} & \texttt{<aug> <slt> csl eu} &  & \texttt{Elevatzaileak hegazkinaren erasotze-angelua...} \\
\hline
\end{tabular}
}
\caption{TSV structure for Augmented SLT.}
\label{tab:aug_slt}
\end{table*}
\vspace{-1ex}
As T5-family models use the padding token as the default start-of-sequence symbol during generation, the \texttt{decoder\_prompt} field in the TSV can be left empty for all tasks. Alternatively, if preferred, it can explicitly be set to \texttt{<pad>}, which results in identical behavior.

Once the TSV files for each individual task and data split have been generated, the rows corresponding to the same split can be concatenated into a single file (e.g., \texttt{train.tsv}, \texttt{validation.tsv}, \texttt{test.tsv}). The resulting files constitute a multitask-compatible dataset with a coherent and standardized structure, ready to be consumed by \emph{MultimodalHugs} without any further preprocessing.

\subsection{Configuration File Creation}

The entire multitask training setup can be specified through a single YAML configuration file, which defines the model architecture, dataset details, processor settings, and training hyperparameters. Below, we outline the key configuration arguments relevant to reproducing the system from \citet{zhang2024scalingsignlanguagetranslation}.

The \textbf{model} section specifies the core architecture and its components. In this setup, we use a T5-family encoder-decoder model as the backbone, initialized with pretrained weights from \texttt{google/byt5-base}. For the multimodal component, following the approach in \citet{zhang2024scalingsignlanguagetranslation}, we embed the pose sequences using a simple linear projection layer. This is reflected by setting the \texttt{multimodal\_mapper\_type} to \texttt{linear}. The relevant configuration is shown below:

\begin{quote}
\small
\texttt{model:} \\
\hspace*{1em}\texttt{type: default\_multimodal\_encoder\_decoder} \\
\hspace*{1em}\texttt{backbone\_type: t5} \\
\hspace*{1em}\texttt{pretrained\_backbone: google/byt5-base} \\
\hspace*{1em}\texttt{multimodal\_mapper\_type: linear} \\
\hspace*{1em}\texttt{\# (additional arguments may be specified as needed)}
\end{quote}

The \textbf{dataset} section defines the paths to the metadata files for each split of the dataset, among other details (\S\ref{subsec:training_setup}). Here, we specify the TSV files previously constructed by concatenating the rows from the different tasks for each split (\texttt{train.tsv}, \texttt{validation.tsv}, and \texttt{test.tsv}). These files conform to the standardized \emph{MultimodalHugs} TSV schema and are ready for ingestion by the framework. The relevant configuration is shown below:

\begin{quote}
\small
\texttt{dataset:} \\
\hspace*{1em}\texttt{train\_metadata\_file: path/to/train.tsv} \\
\hspace*{1em}\texttt{validation\_metadata\_file: path/to/validation.tsv} \\
\hspace*{1em}\texttt{test\_metadata\_file: path/to/test.tsv} \\
\hspace*{1em}\texttt{\# (additional arguments may be specified as needed)}
\end{quote}

The \textbf{processor} section specifies modality-specific preprocessing options. In this case, we set the tokenizer to match the T5-family model (\texttt{google/byt5-base}) and extend its vocabulary with task control tokens (\texttt{<slt>}, \texttt{<agn>}, \texttt{<mt>}, \texttt{<aug>}). Additionally, to reduce the temporal length of pose sequences, we apply subsampling by setting the \texttt{skip\_frames\_stride} parameter to 2, as done in the original work. The relevant configuration is shown below:

\begin{quote}
\small
\texttt{processor:} \\
\hspace*{1em}\texttt{text\_tokenizer\_path: google/byt5-base} \\
\hspace*{1em}\texttt{new\_vocabulary: "<slt>,<agn>,<mt>,<aug>"} \\
\hspace*{1em}\texttt{skip\_frames\_stride: 2} \\
\hspace*{1em}\texttt{\# (additional arguments may be specified as needed)}
\end{quote}

Remaining training hyperparameters, such as learning rate, evaluation strategy, and warmup ratio, should be defined in accordance with the Hugging Face Trainer API and adjusted as needed through the \texttt{training} section of the configuration.

\subsection{Setup and Training}

Once the configuration file is prepared, the complete training pipeline can be launched using the \emph{MultimodalHugs} command-line interface. The first step initializes all training actors—datasets, processors, and model—based on the specified modality and configuration file:
\begin{lstlisting}
    multimodalhugs-setup --modality pose2text --config_path /path/to/config.yaml
\end{lstlisting}

Subsequently, the training process is started by invoking the following command, which executes the sequence-to-sequence training loop and saves outputs to the designated directory:
\begin{lstlisting}
    multimodalhugs-train --task seq2seq --output_path $OUTPUT_PATH [--other_args]
\end{lstlisting}

These two commands suffice to instantiate and execute the entire pipeline without requiring any custom scripts or additional code.

Although we did not run this experiment to completion due to computational constraints, this setup faithfully replicates the architecture and multitask training scheme described by \citet{zhang2024scalingsignlanguagetranslation}, relying solely on standardized TSV metadata and a YAML configuration file.

\clearpage

\section{Example Output from Sign Language Translation}
\label{sec:appendix_predictions}

Below we show a subset of prediction–reference pairs from the SLT experiment (\S~\ref{subsec:slt}), using the model configuration that achieved the best results. Each entry shows the ground-truth label (\texttt{L}) and the corresponding decoded prediction (\texttt{P}), automatically generated and saved by \texttt{multimodalhugs-generate}. The full output file is much longer; here we present an excerpt for illustration:

\begin{lstlisting}[basicstyle=\ttfamily\small,breaklines=true]
L [0]    Hi!
P [0]    Hi, my name is Dean Hale, and I'm a classic teachers of the Austin Tennis Center, here in Austin, Texas.

L [1]    The aileron is the control surface in the wing that is controlled by lateral movement right and left of the stick.
P [1]    The acrylics going to control our strength as well and this is going to help us to move our depths off of the weight and how we can't get any acrylic discourage.

L [2]    By moving the stick, you cause pressure to increase or decrease the angle of attack on that particular raising or lowering the wing.
P [2]    That causes a pressure that's going to get the hip flexors.

L [3]    The elevator is the part that moves with the stick forward and back, and that adjusts the angle of attack of the airplane in the air.
P [3]    So now we're going to have a comb twisted neck and that's how you would use a comb twist to help sort of anchor your comb.

L [4]    Therefore, it's either going uphill, downhill, or flat and that adjusts the air speed that we talked about earlier.
P [4]    If you're looking straight down, the milk will go up to the sky.

L [5]    The rudder is the vertical stabilizer.
P [5]    Now glue is a little bit more like this, and it helps to line up the monitors, to give them more control, so that the neck is rather in line.

L [6]    That's moved by the feet and that's what actually steers the airplane right and left when you start and stop a turn.
P [6]    We're going to use our feet; that's for the tuck.

L [7]    Buenos Dias, I'm Bobby Larew, you didn't know I spoke Spanish, did you?
P [7]    Hi, this is Sean Hobson, and what I'm acentral fish expert in the Austin Tennis Center, here in Austin, Texas.

L [8]    I'm an expert on diving, talking about a back 1 1/2 pike.
P [8]    We're going to talk about the priority of the balloon kick, we're going to talk about the one and the twelve.

L [9]    What I'm going to do with my arms is what's going to help me spin a little faster.
P [9]    You're going to use your arms to make it faster.

L [10]   What you're going to do is you're going to take both your arms and they're going to come underneath your legs.
P [10]   So what I'm going to do is place my blocks underneath my legs.

L [11]   And I'm actually going to lock my wrists when I pike.
P [11]   So what happens is after you get the same strip, you switch it up on your braid.

L [12]   When I lock underneath my legs, I'm going to squeeze in.
P [12]   The benefits of doing your backstroke are that you have to get your knowledgeable.

L [13]   Let me demonstrate you this on my back because it's a lot easier.
P [13]   Now I'm going to show you my back pointer here.

L [14]   This is what it will look like.
P [14]   So you can see I'm doing this pretty tight, and that's what I want to do here.

L [15]   That's a good pike position.
P [15]   That's a good news.
\end{lstlisting}

These decoded predictions are saved automatically and can be used for transparent qualitative analysis or further postprocessing.

\clearpage

\section{Analysis of Hebrew Translation Model Outputs}
\label{appendix:hebrew}

The Hebrew source, English outputs and judgments can be found in: \\ \url{https://docs.google.com/spreadsheets/d/1XY19Oee_FPwp2zw1yjmrjV0fP1WRJPs4Z_DuZyOnNv4}

\subsection{Named Entities}
Our image-based model generally performed better on named entity translation, such as rendering ``Dr. Ehud Ur'' as ``Dr. Ehud O’R,'' which, while not exact, was closer to the reference than the subword-based baseline’s ``Dr. Ahud [Hebrew]'', and still a valid translation. It also correctly produced ``Nova Scotia,'' matching the reference, whereas the baseline mistranslated it as ``Nova Scotland.''. However, the image-based model made serious errors in some cases, such as hallucinating ``Arab-Arab countries'' instead of ``China'' when translating discussions of U.S.-China relations, and inventing nonsensical entities like ``Gittagnet'' instead of ``Gaziantep.''

Conversely, the subword-based baseline excelled in certain cases, such as correctly translating ``ZMapp'' by preserving the spelling, while the image-based model distorted it as ``Zampp.'' It also got ``Gaziantep'' exactly right, unlike the image-based model’s corrupted ``Gittagnet.'' But the baseline often struggled with consistency and sometimes inserted hallucinated details, for example mistranslating ``University of Dalhousie'' as ``University of Delauzzi in Leipzig.''

\subsection{Copying Entities}
The subword-based baseline showed partial success in preserving exact copies of technical terms but frequently introduced small yet critical deviations. For example, it rendered ``802.11n'' with an unnecessary space as ``802.11 n,'' and changed the precise ``2.4GHz'' to ``2.5GHz,'' which alters the technical meaning. It also mistranslated the throughput ``600 Mbit/s'' as ``600 MB/s,'' a major factual error that misstates bandwidth by a factor of eight.

Our image-based model had similar issues, often introducing more severe distortions. It hallucinated ``802.111'' instead of ``802.11n,'' completely breaking the standard’s identifier, and expanded ``2.5GHz'' to ``5.8GHz,'' further diverging from the original spec. Additionally, it produced nonsensical units like ``600/GB'' for throughput. These consistent failures in both systems highlight a systemic weakness in handling exact numerical and technical terms, which require literal copying rather than translation or normalization.

\subsection{Other}

Beyond named entities and exact copies, our image-based model and the subword-based baseline exhibited distinct general behaviors. The image-based model tended to produce translations with greater fluency and accuracy at the sentence level, generating smoother and more natural English phrasing in many cases. For example, our image-based model’s ``starring Ryan Gosling'' accurately reflected the intended meaning compared to the baseline’s incorrect ``directed by Ryan Gosling.'' However, the image-based model often introduced hallucinations—fabricating content or significantly altering details not present in the source—particularly with less common words or rare morphological forms. This reflects the model's tendency to overgeneralize from visual patterns, which, while beneficial for correcting misspellings and capturing named entities, can also lead to major semantic errors.

In contrast, the subword-based baseline was more literal and conservative, usually sticking closely to the source structure and preserving word order even when it resulted in awkward or ungrammatical English. While this conservative approach reduced hallucinations, it often produced stilted or incomplete sentences and struggled with handling Hebrew's rich morphology—frequently fragmenting words into subtokens that lost or confused meaning. This difference underscores a trade-off: the image-based approach improves handling of word-level noise and named entities at the cost of occasional severe hallucinations, while the subword-based baseline maintains higher fidelity to the source syntax but produces less fluent, sometimes unnatural outputs.

\clearpage

\section{Future Training Workflow Example: Mixed-Modality Sequences}
\label{appendix:future-workflow}

\begin{figure*}[ht!]
    \centering
    \includegraphics[width=\textwidth]{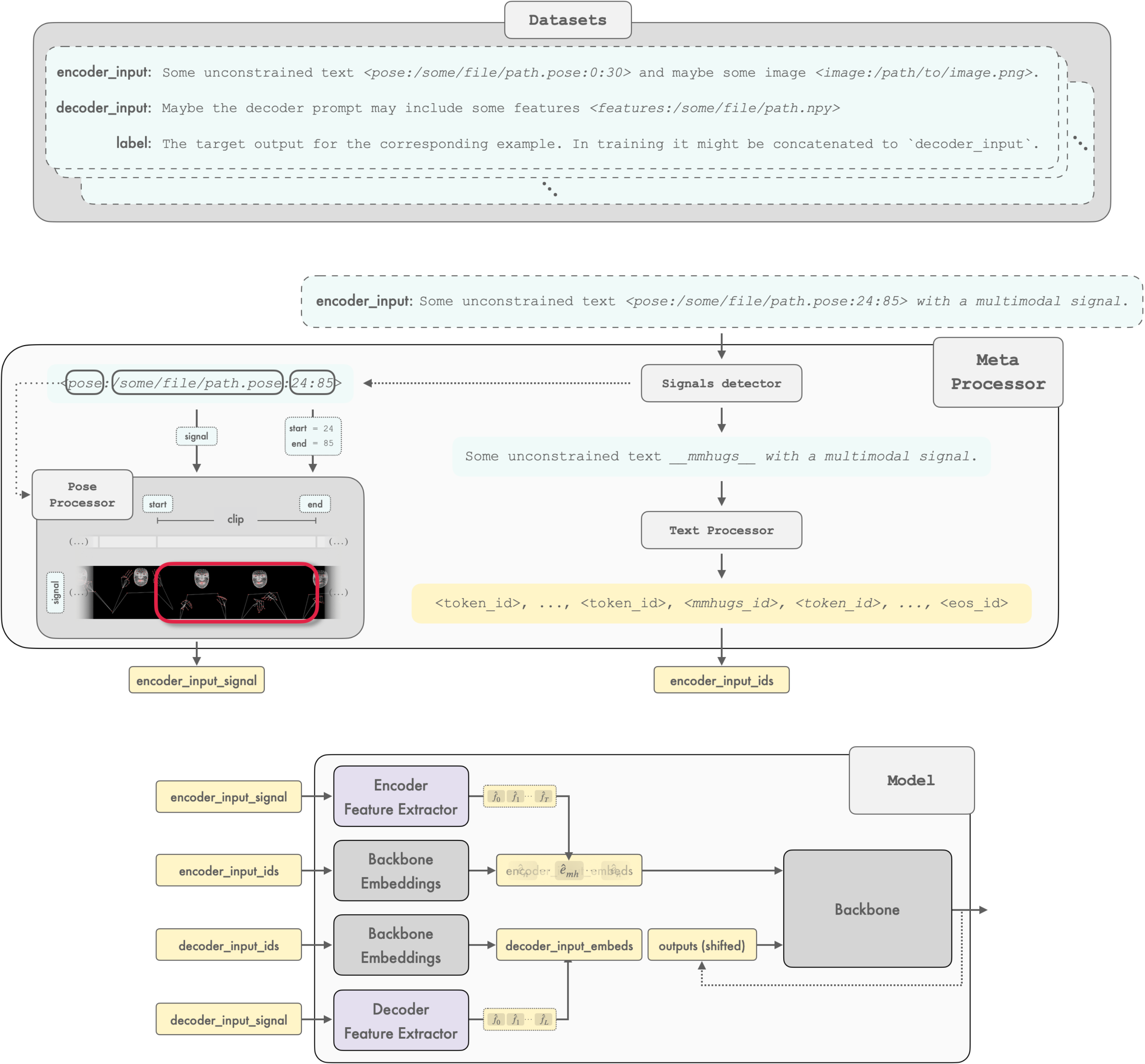}
    \caption{Envisioned future training pipeline for mixed-modality sequences in \emph{MultimodalHugs}. The dataset fields contain unconstrained text interleaved with references to multimodal resources. The Meta Processor detects, parses, and routes each signal to the corresponding modality-specific processor, while passing remaining text through the Text Processor. Resulting multimodal features and token IDs are then aligned and fed into the model for training.}
    \label{fig:multimodalhugs_future_training_workflow}
\end{figure*}

Figure~\ref{fig:multimodalhugs_future_training_workflow} illustrates our envisioned training pipeline for mixed-modality input sequences, as described in (\S\ref{sec:conclusion_and_future_work}). This future workflow builds upon the current design of \emph{MultimodalHugs} (see Appendix~\ref{appendix:pose-workflow}) but introduces a more flexible mechanism that enables users to freely interleave textual and non-textual signals in a single sequence. 

The central idea is to allow unconstrained input text—containing inline references to multimodal resources (e.g., pose files, images, audio clips)—to be parsed and aligned automatically at runtime. The dataset for each example includes three fields: \texttt{encoder\_input}, which is a natural-language sequence with embedded signal references; an optional \texttt{decoder\_input}, which may include textual prompts or features; and the \texttt{label}, which serves as the output target.

The workflow use a \textit{Meta Processor}, which detects multimodal signals within input and output sequences using a dedicated signals detector component. These detected signals are then routed to appropriate modality-specific processors (e.g., the Pose Processor in the example), which extract the corresponding features while respecting temporal markers or clip boundaries. Meanwhile, the remaining textual portion of the input is passed through the Text Processor to produce tokenized sequences.

The resulting outputs are combined to produce multiple synchronized streams of encoder inputs: token IDs for the text portions and embedded multimodal features for the detected signals. These streams are then fed into the model, where feature extractors and embedding layers handle both textual and non-textual modalities. The decoder receives its own tokenized inputs and embeddings and generates the output sequence as usual.

This flexible workflow enables seamless mixing of modalities within the same sequence, giving users full control over the composition and placement of multimodal signals in their inputs. It eliminates the need for rigid input templates or hardcoded signal placement, aligning with trends in decoder-only, multimodal language models.

\clearpage

\twocolumn



\end{document}